\documentclass[a4paper, 10pt, conference]{IEEEtran}  
\IEEEoverridecommandlockouts
\usepackage[pdftex]{graphicx}
\def\BibTeX{{\rm B\kern-.05em{\sc i\kern-.025em b}\kern-.08em
    T\kern-.1667em\lower.7ex\hbox{E}\kern-.125emX}}
\usepackage{cite}
\usepackage{textcomp}
\usepackage{comment}

\usepackage{lmodern}

\usepackage{amsmath}
\usepackage[psamsfonts]{amssymb}
\usepackage{amsfonts}
\usepackage{braket}
\usepackage{latexsym}

\usepackage{multirow}

\begin{document}




\newcommand{\oomoji}[1]{{\mbox{\boldmath $#1$}}}
\newcommand{\tenti}{^{\rm T}}

\title{Difficulty in estimating visual information from randomly sampled images}

\author{\IEEEauthorblockN{1\textsuperscript{st} Masaki Kitayama}
\IEEEauthorblockA{\textit{Tokyo Metropolitan University}, \\ Tokyo, Japan \\
}
\and
\IEEEauthorblockN{2\textsuperscript{nd} Nobutaka Ono}
\IEEEauthorblockA{\textit{Tokyo Metropolitan University}, \\ Tokyo, Japan \\
}
\and
\IEEEauthorblockN{3\textsuperscript{rd} Hitoshi Kiya}
\IEEEauthorblockA{\textit{Tokyo Metropolitan University}, \\ Tokyo, Japan \\
}
}

\maketitle

\begin{abstract}
In this paper, we evaluate dimensionality reduction methods in terms of difficulty in estimating visual information on original images from dimensionally reduced ones.
Recently, dimensionality reduction has been receiving attention as the process of not only reducing the number of random variables, but also protecting visual information for privacy-preserving machine learning.
For such a reason, difficulty in estimating visual information is discussed. In particular, the random sampling method that was proposed for privacy-preserving machine learning, is compared with typical dimensionality reduction methods.
In an image classification experiment, the random sampling method is demonstrated not only to have high difficulty, but also to be comparable to other dimensionality reduction methods, while maintaining the property of spatial information invariant.

\end{abstract}

\section{Introduction}
Recently, it has been very popular to utilize cloud servers to carry out machine learning algorithms instead of using local servers. However, since cloud servers are semi-trusted, private data, such as personal information and medical records, may be revealed in cloud computing. For the reason, privacy-preserving machine learning has become an urgent challenge\cite{homomorphic1, 2007.08775, sirichotedumrong2019privacy, sirichotedumrong2019pixel, 2006.01342}.
In this paper, we focus on dimensionality reduction methods in terms of two issues: difficulty in estimating visual information on original images from dimensionally reduced ones, and performance that reduced data can maintain in an image classification experiment.
In machine learning, dimensionality reduction is used for not only reducing the number of random variables, but also protecting visual information for privacy-preserving machine learning. However, dimensionality reduction methods have never been evaluated in terms of above the two issues at same time.

For such a reason, difficulty in estimating visual information is discussed. In particular, the random sampling method that was proposed for privacy-preserving machine learning\cite{aprilpyone2020encryption}, is compared with typical dimensionality reduction methods such as random projection and PCA\cite{bingham2001random, wold1987principal}.
In an image classification experiment, the random sampling method is demonstrated not only to maintain high difficulty, but also to have close machine learning performance to that of the random projection method.

\section{Linear Dimensionality Reductions}
Let us consider a projection from a vector $x\in\mathbb{R}^D$ to a low-dimensional vector $y\in\mathbb{R}^K (K<D)$.
If the projection can be represented by using a matrix $P\in\mathbb{R}^{K\times D}$ as

\begin{align}
y=Px\quad, \label{linear dim reduction}
\end{align}
it is a linear dimensionality reduction and $P$ is called a projection matrix.
In machine learning, $P$ is used for reducing the number of random variables for avoiding negative effects of high-dimensional data.
The random projection method\cite{bingham2001random} and principal component analysis (PCA) are typical linear dimensionality reduction methods.
The random projection is a method that does not use any statistics of dataset, but PCA is not.
For the random projection, elements of $P$ have a normal distribution with an average value of 0 and a variance of $\sqrt{1/K}$. Therefore, the random projection is not required to calculate any statistics of dataset for designing a projection matrix $P$.

\section{Random Sampling}
The random sampling method was proposed as a dimensionality reduction method for privacy preserving machine learning\cite{randomsampling}.
It is also expected to be applied to deep convolutional neural network, due to the property of spatial information invariant\cite{aprilpyone2020encryption}.

Let us consicer applying the random sampling method to a pixel vector $x\in\mathbb{R}^D$ of an image to create $y\in\mathbb{R}^K (K<D)$.
Next, let $\{\phi(i)\, |\, i=1,\ldots,K\}$ denote $K$ indexes selected from $D$ pixel indexes,
where $\phi(i)\neq\phi(i^\prime)$ if $i\neq i^\prime$, randomly generated with a seed.
By using $\phi(i)$, the random sampling operation can be written as

\footnotesize
\begin{align}
\label{random sampling}
y=(x_{\phi(1)},x_{\phi(2)},\ldots,x_{\phi(K)})^{\rm T}\quad,
\end{align}
\normalsize
where $x_{\phi(i)}$ is the $\phi(i)$-th element of $x$.
Here, if we define a matrix $P\in\mathbb{R}^{K\times D}$ with elements $p_{ij} (i=1,\ldots,K,j=1,\ldots,D)$ defined by
\footnotesize
\begin{align}
& p_{i,j}=\left\{
\begin{array}{ll}
1 & (j=\phi(i)) \\
0 & ({\rm otherwise})
\end{array}
\right. \quad, \label{random sampling 2}
\end{align}
\normalsize
the random sampling is reduced to the form of Eq.(\ref{linear dim reduction}).
That is, the random sampling is a linear dimensionality reduction, and is a method that does not use the statistics of dataset as well as the random projection.

\section{Visual Information Estimation}
\label{attacks}
Assuming that an attacker knows a projection matrix $P$ used in dimensionality reduction, difficulty in estimating visual information on plain images is discussed.
The attacker's goal is to create $Q$ to approximatelly reconstruct the target image $x$ from the low dimensional vector $y$ as
\begin{align}
x^\prime=Qy\quad. \label{reconstruction}
\end{align}
In this paper, two attacks are considered  to estimate $Q$.

\subsection{Attack With Pseudo-Inverse Matrix}
An attacker can use a pseudo-inverse matrix ($Q_{\rm pinv}$) of projection matrix $P$ to estimate visual information on original images,
where $Q_{\rm pinv}$ is designed by using an algorithm with the singular-value decomposition of $P$\cite{harville1998matrix}.

\subsection{Regression Attack With Attacker's Dataset}
\label{regression attack}
An attacker first prepares his own dataset ($X_{\rm attack}$) and a dataset ($Y_{\rm attack}$) projected from $X_{\rm attack}$ by using $P$, and then designs a linear reconstruction matrix ($Q_{\rm reg}$) that regresses $X_{\rm attack}$ from $Y_{\rm attack}$ in accordance with the least squares method.
In general, the effectiveness of this attack depends on the relation between the distribution of $X_{\rm attack}$ and that of target images.
Therefore, in this paper, we classify $X_{\rm attack}$ into two types in accordance with the distribution of $X_{\rm attack}$.
\begin{itemize}
\item type 1: $X_{\rm attack}$ consists of images with the same class-labels and distribution as those of the target images.
\item type 2: $X_{\rm attack}$ consists of images with class-labels and a distribution that are different from those of the target images.
\end{itemize}

\begin{figure}[t]
  \begin{center}
   \includegraphics[scale=0.47]{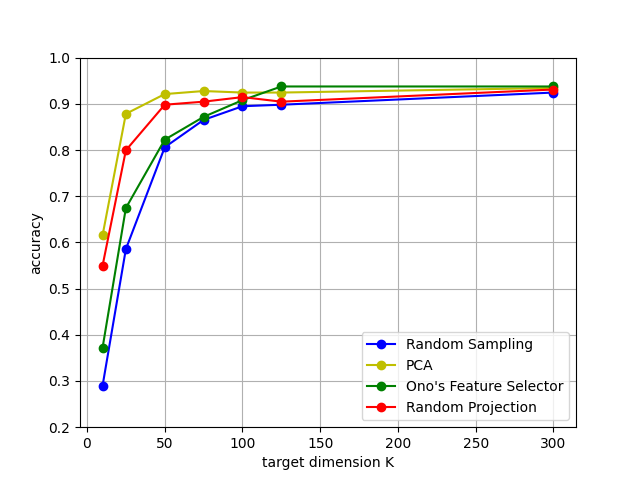}
   \caption{Classification accuracy with various dimensionality reduction methods (Linear SVM). \label{accs linsvm}} 
  \end{center}
\end{figure}

\begin{figure}[t]
  \begin{center}
   \includegraphics[scale=0.47]{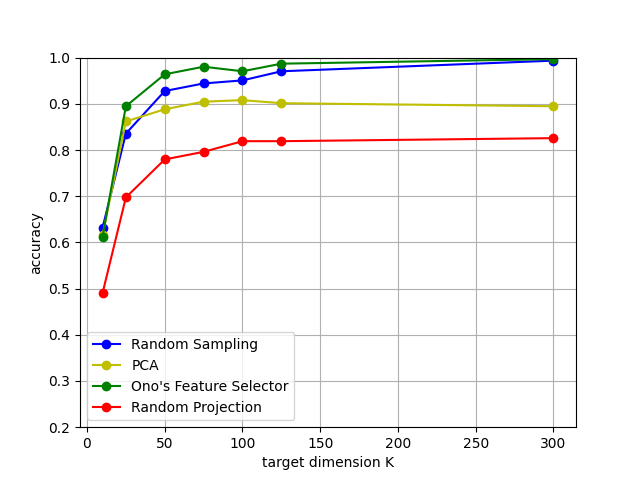}
   \caption{Classification accuracy with various dimensionality reduction methods (Random Forest). \label{accs rf}} 
  \end{center}
\end{figure}

\section{Experiment}
Face-image classification experiments were carried out for evaluating the random sampling method in terms of both classification accuracy and difficulty in estimating visual information.
The dataset was Extended Yale Database B\cite{Yale}, which contains 38 individuals and 64 frontal facial images with 168$\times$192 pixels per each person.
Each image was normalized to a size of 28$\times$28, so it had $D=784$ dimension as a vector,
we splitted the dataset into two datasets: $X_{\rm main}$ and $X_{\rm sub}$, each of which had 19 classes and 1216 images, without duplication of classes.
Moreover, $X_{\rm main}$ was divided into $X_{\rm train}$ (912 images) for training and $X_{\rm test}$ (304 images) for testing.

We also used the CIFAR-10\cite{krizhevsky2009learning} dataset: $X_{\rm CIFAR-10}$ for evaluating difficulty in estimating visual information on $X_{\rm test}$.
This dataset consists of 60k images with 10 classes such as dogs and ships,
whose distribution is different from $X_{\rm train}$, $X_{\rm test}$ and $X_{\rm sub}$.

Finally, each vector $x\in\{X_{\rm train}$, $X_{\rm test}$, $X_{\rm sub}$, $X_{\rm CIFAR-10}\}$ was projected to $y\in\{Y_{\rm train}$, $Y_{\rm test}$, $Y_{\rm sub}$, $Y_{\rm CIFAR-10}\}$ with a target dimension ($K$) by using the random sampling and three dimensionality reduction methods: the random projection, PCA, and a feature selection algorithm proposed by 
Ono\cite{featureselector}.
PCA and Ono's method require calculating the statistics of $X_{\rm train}$, but the random sampling and random projection do not.

\begin{figure*}[t]
  \begin{center}
   \includegraphics[scale=0.77]{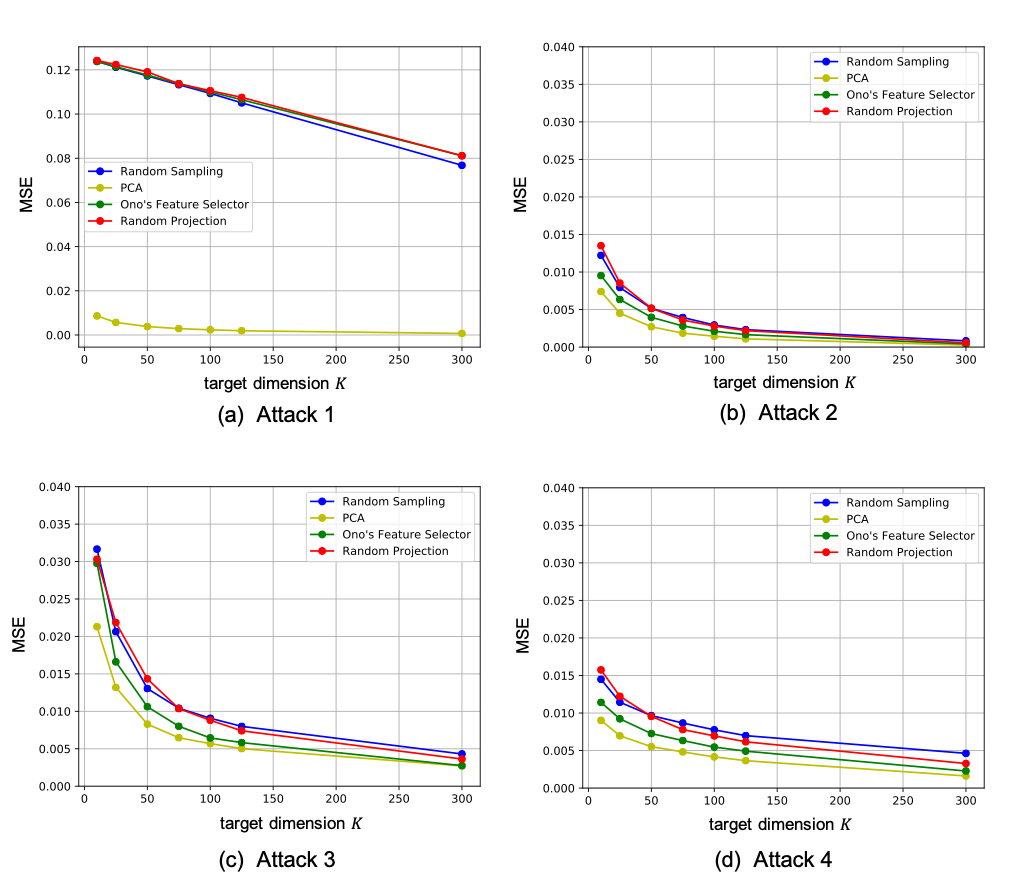}
   \caption{Mean squared error between original image and reconstructed image.  \label{mse}}
  \end{center}
\end{figure*}

\subsection{Machine Learning Performance}
We trained a random forest classifier and SVM with the linear kernel by using $Y_{\rm train}$, and tested by using $Y_{\rm test}$.
Figures \ref{accs linsvm} and \ref{accs rf} show the comparison of the dimensionality reduction methods in term of classification accuracy.
Under the use of SVM, the random sampling had a similar performance to Ono's method.
For the random forest, the random sampling also has almost the same accuracy as that of Ono's method.
As a result, the random sampling was demonstrated to be comparable with other dimensionality reduction methods, while maintaining the property of spatial-information invariant.

\subsection{Robustness Against Visual Information Estimation}
Assuming that an attacker knows only projection matrix $P$ used in dimensionality reduction, difficulty in estimating visual information on $X_{\rm test}$ was evaluated.
We assumed the following four attacks to estimate $X_{\rm test}$ from $Y_{\rm test}$.
\begin{itemize}
\item Attack 1: Attack using a pseudo-inverse matrix of $P$.
\item Attack 2: Regression attack with $X_{\rm train}$ (type 1 in section \ref{regression attack}).
\item Attack 3: Regression attack with $X_{\rm CIFAR-10}$ (type 2 in section \ref{regression attack}).
\item Attack 4: Regression attack with $X_{\rm sub}$.
\end{itemize}

In attack 4, the attacker does not have facial images of the people included in $X_{\rm test}$, but knows the conditions under which  $X_{\rm test}$ was taken.

Figure \ref{mse} shows MSE values between $X_{\rm test}$ and $X_{\rm test}^\prime$.
From the figure, the random sampling was relatively robust compared with the other dimensionality reduction methods.
The absolute values of MSE of the random sampling were large for attacks 1 and 3,
but were small for attacks 2 and 4 as well as the other methods.

We defined accuracy reduction ratio (ARR) as another criterion.
First, we trained a logistic regression classifier ($\theta$) by using $X_{\rm train}$, and then the ARR is defined as
\begin{align}
{\rm ARR}=\frac{{\rm ACC}_\theta(X_{\rm test})-{\rm ACC}_\theta(X_{\rm test}^\prime)}{{\rm ACC}_\theta(X_{\rm test})}\quad, \label{definied arr}
\end{align}
where ${\rm ACC}_\theta(X)$ is the function which returns the accuracy when dataset $X$ is applied to $\theta$.
A high ARR value indicates that $X_{\rm test}^\prime$ has low class-specific visual information, i.e., the dimensionality reduction is robust against the attack.

\begin{figure}[t]
  \begin{center}
   \includegraphics[scale=0.68]{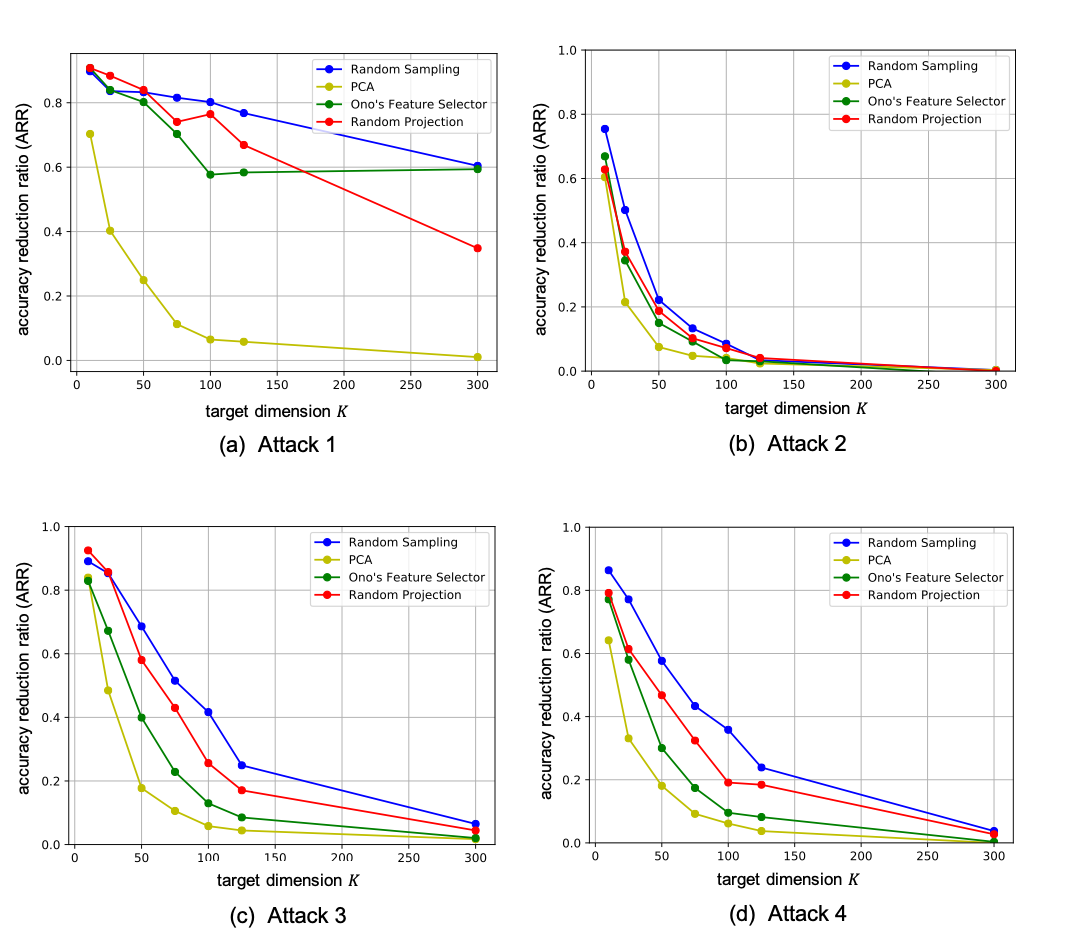}
   \caption{Comparison of accuracy reduction ratio values.  \label{arr}}
  \end{center}
\end{figure}

Figure \ref{arr} shows the comparison of ARR values.
From the figure, the random sampling was demonstrated to be robust against attack 4.
It indicates that attack 4 did not effectively recover the class-specific information of the target images.

\begin{figure}[b]
  \begin{center}
   \includegraphics[scale=0.7]{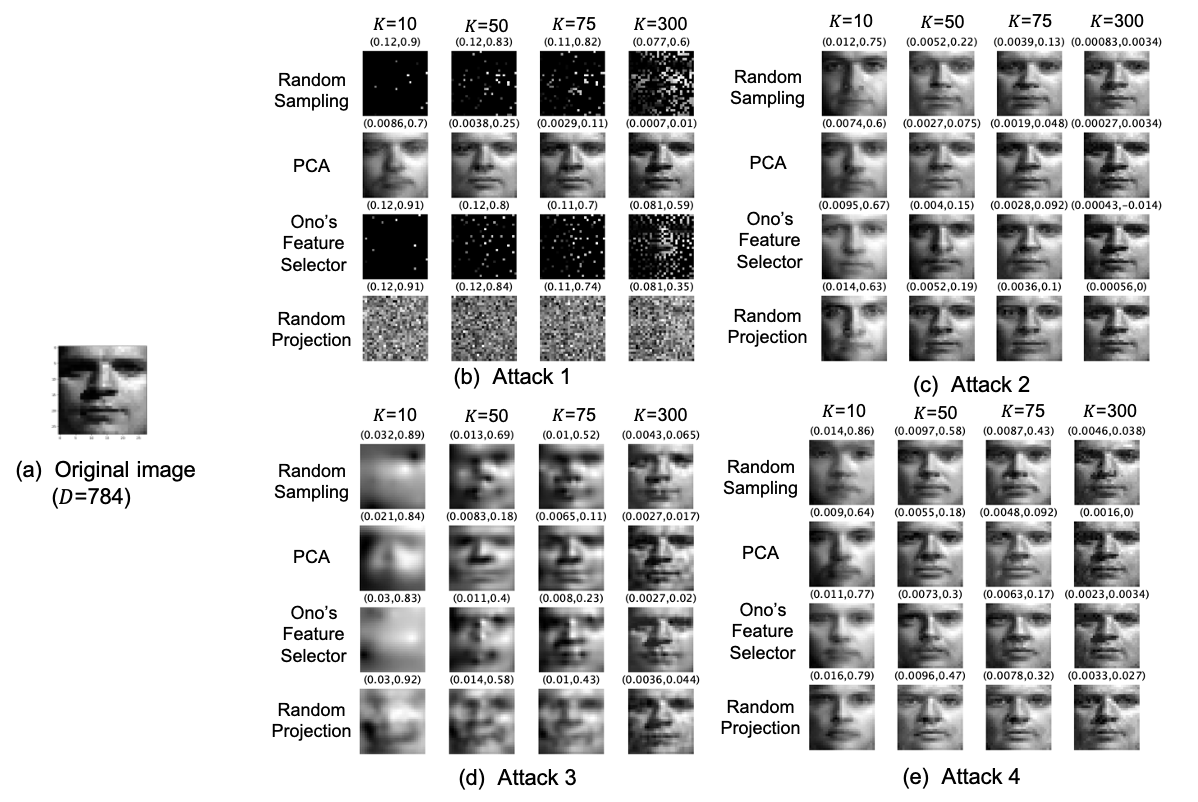}
   \caption{Examples of reconstructed images ($D$=784) with MSE (left) and ARR (right) values.   \label{reconst images}}
  \end{center}
\end{figure}

Figure \ref{reconst images} shows an original images and examples of reconstructed images.
Although the images reconstructed by attack 4 can be easily interpreted as human faces,
the facial features were different from the original person.

\onecolumn

\twocolumn

\section{conclusion}
In this paper, we compared the random sampling method with typical linear dimensionality reduction methods in terms of both machine learning performance and difficulty in estimating visual information in face classification experiments. The random sampling was demonstrated not only to have the property of spatial position invariant which is useful for privacy-preserving learning, but also to maintain comparable performance to other dimensionality reduction methods and difficulty in visual information estimation under practical conditions.

\bibliographystyle{IEEEbib}
\bibliography{ref_kitayama20190325}

\end{document}